\documentclass[letterpaper, 10 pt, journal, twoside]{IEEEtran}  

\usepackage{graphicx}
\usepackage{epsfig}
\usepackage{bm}
\usepackage{amsmath}

\begin{document}
\title{
Mamba as a motion encoder\\ for robotic imitation learning
}

\author{Toshiaki Tsuji$^{1}$
\thanks{$^{1}$Toshiaki Tsuji is with the 
Graduate School of Science and Engineering, Saitama University, 255 Shimo-okubo, Saitama, 338-8570 
{\tt\small E-mail: tsuji@ees.saitama-u.ac.jp}}%
}%

\maketitle

\begin{abstract}
Recent advancements in imitation learning, particularly with the integration of LLM techniques, 
are set to significantly improve robots' dexterity and adaptability. 
This paper proposes using Mamba, a state-of-the-art architecture with potential applications 
in LLMs, for robotic imitation learning, highlighting its ability to function as an encoder that 
effectively captures contextual information.
By reducing the dimensionality of the state space, Mamba operates similarly to an autoencoder.  
It effectively compresses the sequential information into state variables while preserving the essential 
temporal dynamics necessary for accurate motion prediction.
Experimental results in tasks such as Cup Placing and Case Loading demonstrate that 
despite exhibiting higher estimation errors, Mamba achieves superior success rates compared to 
Transformers in practical task execution. 
This performance is attributed to Mamba’s structure, which encompasses the state space model. 
Additionally, the study investigates Mamba’s capacity to serve as a real-time motion generator 
with a limited amount of training data.
\end{abstract}

\section{Introduction}
In the field of robotic manipulation, the application of imitation learning has been rapidly advancing in 
recent years~\cite{osa2018algorithmic, ravichandar2020recent}.
Conventional robot control methods relying on manual coding necessitate complex motion planning 
as well as accurate modeling of the environment and objects,
imitation learning offers the advantage of intuitively and efficiently teaching tasks to robots.
Furthermore, with the recent advancements in deep learning techniques, it is becoming possible to acquire 
generalized control policies adaptable to various environments in robotics.
So far, various architectures have been proposed, such as behavior cloning~\cite{pomerleau1988alvinn, ross2011reduction},
inverse reinforcement learning~\cite{russell1998learning}, and GAIL~\cite{ho2016generative}.
There is an increase in multimodal imitation learning that extends modalities~\cite{su2016learning, noda2014multimodal}
and models representing complex tasks using hierarchical architecture~\cite{bentivegna2004learning, takeuchi2023motion} has a potential to extend imitation learning to complicated tasks. 
Integration of force control and imitation learning is essential for enhancing performance of contact rich 
tasks~\cite{abu2020variable, sasagawa2020imitation}. 
Depending on the development of these imitation learning studies, 
there is potential for robots to attain human-level dexterity 
and adaptability even in tasks that have been considered difficult to automate.

This research aims to verify the significance of using deep state-space 
models (SSMs)~\cite{hippo, s4, mamba}, particularly Mamba~\cite{mamba},
as a model for robotic imitation learning.
The rapid development of LLMs has not only impacted language tasks but also influenced fields 
such as image and speech processing, drawing attention to their potential applications in robotics~\cite{xu2024survey}.
In robotic imitation learning, LLMs are expected to contribute to mapping language instructions to physical 
actions~\cite{mees2022matters, ahn2022can, shridhar2023perceiver}. 
Not only that, but they have also been applied to action generation and path planning due to their versatile knowledge 
representation and multimodal integration capabilities~\cite{chaplot2021differentiable, johnson2021motion, kobayashi2024ilbit}.  
While many of these studies are based on Transformers with attention mechanisms,
Mamba is also a good candidate as a model that achieves equivalent long-term memory retention. 
As Mamba is based on a continuous SSM, it could exhibit better characteristics than Transformers
as a model for robotic imitation learning, which deals with sequences of continuous sensor-actuator responses. 
Addressing the challenges of limited training data and the need for quick responses, 
this research develops a model with simple configuration suitable for real-time motion generation in robots. 
The capability of the simplified model is examined to assess its potential as a motion generation model for IL. 
Previous studies have been designed on the premise of incorporating long sequences of states entirely into the model,
but to achieve both real-time responsiveness and adaptability to diverse environments, it is desirable for designers to be
able to handle and process past context and current state information separately.
Additionally, for data storage and real-time processing, it is preferable that each piece of information is appropriately compressed.
Mamba’s ability to function as an encoder that stores context as low-dimensional information is leveraged, 
extending it into a model that assigns past contextual information to state variables. 

\section{Related studies}
\subsubsection*{Imitation Learning}
Early imitation learning methods primarily focused on learning one-to-one mappings between state-action pairs.
Despite showing promising results in tasks such as autonomous driving~\cite{pomerleau1988alvinn} and robot control~\cite{ross2011reduction, florence2022implicit},
these methods overlooked the rich temporal information contained in histories.
Therefore, numerous studies have successfully incorporated action sequences using RNNs~\cite{beliaev2022imitation, kutsuzawa2019trajectory}.
LSTMs~\cite{hochreiter1997long} have been widely applied in many tasks where long-term dependencies need to be 
considered.
Seq2Seq models adopting LSTM as both encoder and decoder allow for optimization of generated trajectories in latent space,
enabling dimensionality reduction of optimization problems, as their latent variables compress sequence information~\cite{kutsuzawa2019trajectory}.
However, limitations in memory retention have been pointed out as potentially problematic when applied to very long sequences~\cite{zhou2021informer}.
Transformers are widely applied in recent imitation learning due to their ability to model longer sequences
than LSTM while maintaining training efficiency through parallelization of sequence processing.
Having developed in language and image processing, multimodal imitation learning using Transformers to encode
both image and language sequences is being researched~\cite{lu2019vilbert,hu2021unit} and there are also 
studies in robotic applications, too~\cite{jiang2023vima,shah2023mutex}.
As mentioned above, the Transformer is a powerful tool for robot imitation learning due to its high memory retention performance. 
This paper proposes a model that achieves memory retention on par with the Transformer, making it suitable for enhancing robot task performance. 

\subsubsection*{SSM}
SSMs are also promising candidates to solve the issue in robotic imitation learning.
SSMs have long been applied to time series information prediction~\cite{kitagawa1996monte, durbin2000time}.
Recently, there have been many attempts to combine SSMs with deep learning.
Watter {\it et al}. proposed an SSM combining locally linear state transition models
with nonlinear observation models using VAE~\cite{watter}.
Gu {\it et al}. proposed HIPPO, a model applying SSMs to discrete neural networks~\cite{hippo},
and extended it to S4~\cite{s4} and Mamba~\cite{mamba}.
These can achieve efficiency and performance surpassing Transformers in processing long-range dependencies.
It has long been known that CNNs are effective in extracting features necessary for motion generation from environmental information,
and as Mamba is reported to possess these characteristics, it is promising as an IL policy for robots,
with examples of application to robot imitation learning already emerging~\cite{robomamba, jia2024mail}.
Jia {\it et al}. demonstrated a method to extend Mamba to an encoder-decoder variant by augmenting input with learnable action,
state, and time embedding variables, enabling efficient learning and applying it to diffusion processes~\cite{jia2024mail}.
Previous researches applying Mamba to robots have all been designed assuming complex contexts with large amounts of data,
resulting in large network structures with stacked layers of residual mechanisms.
This paper focuses on the fact that the context of robot manipulation tasks can be often represented by smaller number of words
compared to general sentences, and reduces the dimensionality of the state space. 
To accommodate learning with a small amount of training data, the number of Mamba layers is reduced, resulting in a simplified, 
low-dimensional configuration. The capability of this configuration as an IL motion generation model is then evaluated.
Additionally, the dimensionality reduction of the state space means that the amount of contextual information can be reduced.
Utilizing this feature, the model is improved to generate actions using only the compressed contextual information and recent
input sequences.

\subsubsection*{Autoencoder}
The model adopted in this research can be interpreted as utilizing Mamba as an autoencoder, given its low-dimensional SSM configuration. 
This approach bears relevance to previous works employing autoencoders in robotics, such as \cite{noda2014multimodal, kutsuzawa2019trajectory, zambelli2020multimodal, meo2021multimodal}. 
Some of these studies also have architectures that provide inputs to latent variables~\cite{kutsuzawa2019trajectory}. 
The most significant difference of the present study compared to these is the CNN and the selection mechanism. 
These features are expected to enable selective extraction of crucial information while mitigating performance degradation in long sequences.

\section{Proposed method}
\subsection{Architecture of State Space Model}
\begin{figure}[tb]
  \centering
  \includegraphics[width=6.2cm]{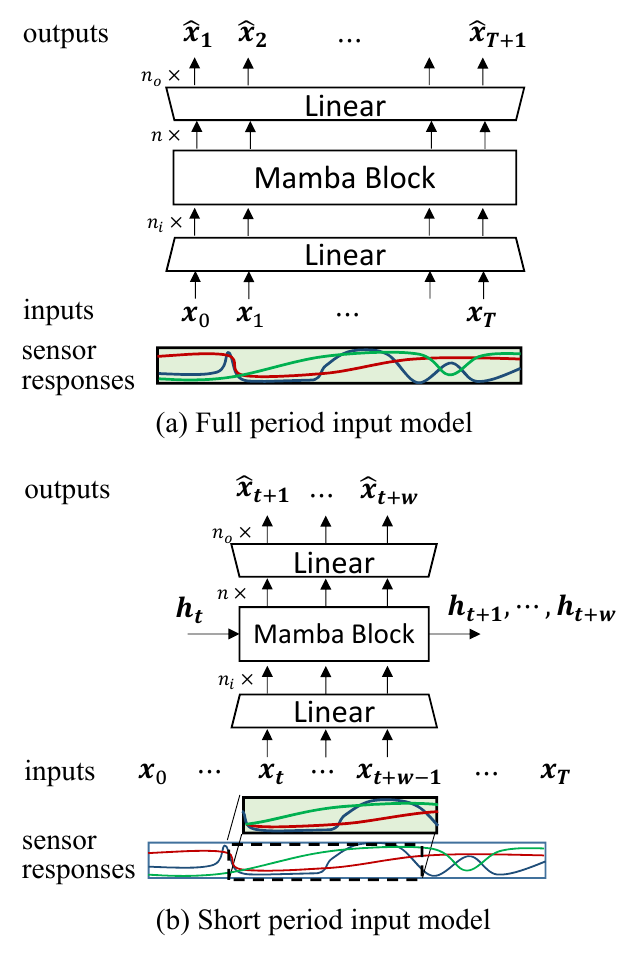}
  \caption{
    Architecture of the proposed model.}
  \label{fig:model}
\end{figure}

\begin{figure}[tb]
  \centering
  \includegraphics[width=6.2cm]{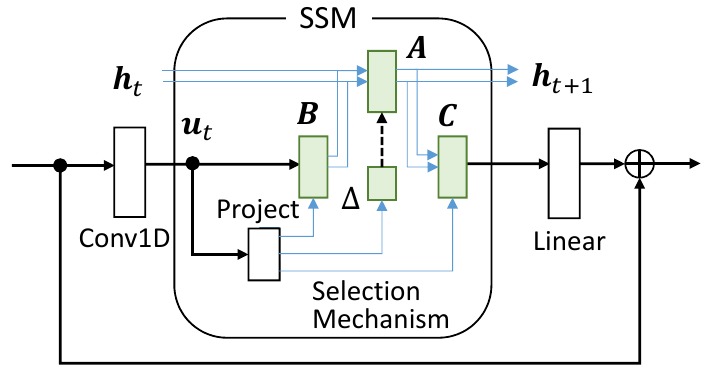}
  \caption{Mamba block.}
  \label{fig:mambablock}
\end{figure}

The machine learning model implemented in this research is shown in Fig.~\ref{fig:model}(a). 
Here, $\bm{x}_t$ represents the input vector at time $t$, the hat symbol denotes its predicted value, 
and $T$ represents the length of the sequence.
Mamba traditionally inputs sequence information vectorized by a tokenizer into multiple overlapping Mamba layers, 
and then converts it back to words for output. In this research, the tokenizer is replaced with a linear layer 
because continuous data is given for input.
Additionally, the input dimension to the Mamba block is reduced by this linear layer. The computation results of the low-dimensional Mamba block are then increased to the same dimension as the input through a linear layer for output.
To predict future trajectories, the output time is advanced compared to the input. Also, to avoid performance degradation due to padding in the Convolution layer, the input layer is made longer by the kernel length. While not strictly identical, the input and output are nearly equivalent to an autoencoder structure.
The configuration of the Mamba block is shown in Fig.~\ref{fig:mambablock}. 

It features a residual network with skip connections that perform identity mapping 
(see the bottom arrow in Fig.~\ref{fig:mambablock}), similar to the model in \cite{mamba}, 
but modified for application in robot trajectory generation. 
Previous studies applying Mamba to robots \cite{robomamba, jia2024mail} have made Mamba blocks 
handle features enhanced by encoders or signal processing processes on input and output data. 
In contrast, this research adopts a simpler model configuration consisting solely of Mamba blocks.

The input consists of 16-axis time series data: 
8-axis angles and 8-axis torques corresponding to joint motors of the robot. 
The internal structure of the residual network consists of convolutional layers, SSM, and linear layers. 
The state variable $\bm{h}_t+1$ and output $\bm{y}_t+1$ of SSM at time $t+1$ are derived using the 
following equations.
\begin{eqnarray}
\left\{
\begin{array}{l}
\bm{h}_{t+1} = \bar{\bm{A}}\bm{h}_{t} + \bar{\bm{B}}\bm{u}_{t} \\
\bm{y}_{t+1} = \bm{C}\bm{h}_{t+1} \
\end{array}
\right.
\end{eqnarray}
\begin{eqnarray}
\bar{\bm{A}} &=& \exp (\bm{\Delta} \bm{A}), \label{eq:abar}\\
\bar{\bm{B}} &=& (\bm{\Delta A})^{-1}(\exp (\bm{\Delta A}) - \bm{I}) \cdot \bm{\Delta B}. 
\end{eqnarray}
In these equations, $\bm{A}$, $\bm{B}$, and $\bm{C}$ represent the state transition matrix, input matrix, 
and output matrix, respectively, with the superscript bars indicating the parameters of the discretized equations. 
$\bm{u}$ is the input to SSM. 
The parameters $\bm{B}$, $\bm{C}$, and $\bm{\Delta}$ are variable and adjusted by a feed-forward neural network through a Selection Mechanism. 
%
\subsection{Design of the proposed method}
The configuration of SSM is based on \cite{mamba}, while the dimensionality of the SSM is reduced 
based on the assumption that the variation in robot movements is limited compared to the highly diverse patterns in natural language. 
In this study, the SSM dimensionality was set to 4, achieving characteristics similar to an autoencoder.
In addition, this study shows that the performance of motion generation is sufficient even in an extremely simplified configuration. 
The number of input and output linear layers $n_i$, $n_o$ and the number of Mamba blocks $n$  are 1, 3, and 1, respectively. 
Dropout layers with a retention probability of 0.75 were incorporated between the linear layers to regularize 
the network and prevent overfitting.

While Mamba implements functionality akin to the Attention mechanism 
by incorporating a gated MLP (gMLP) within its block,  
the current study adopts a structure that excludes gMLP when the dimension of SSM is low 
due to its tendency to cause overfitting and unstable learning. 
When performing advanced tasks that require more complex context, 
including gMLP may improve performance. 
It is desirable to use them selectively depending on the application.
Most of the continuous-time theories in control and signal processing are based on the assumption that all past responses influence 
the present, so this modification is not unreasonable for imitation learning in robotics. 

The residual network outputs the estimated values of 16-axis position and force data $N$ samples ahead. 
In this study, the angles and torques of each joint are divided into 8 axes each. 
During training, the time series data of position and force are inputted for each sample, 
and the estimated position and force data are compared with the post-sample training data to obtain the error. 
The loss is calculated from this error, and the model is trained through backpropagation.

Mamba has convolutional layers close to the input, and SSM recursively uses the low-dimensional information of 
past histories as input for the next step, thus exhibiting features of both CNN and RNN. 
Also, previous studies have provided the entire input sequence at once, while the model in this paper, 
as shown in Fig.~\ref{fig:model}(b), is structured to provide the input sequence only for a certain period. 
Here, $w$ is the length of the short period input. 
State variables generated in past steps are input to reflect the context before the input period in motion generation. 
Since state variables are low-dimensional compared to input, the amount of information to be preserved is reduced.

It is often a problem in RNNs that their performance degrades in tasks that require consideration of long-term dependencies. 
This challenge can be overcome by setting the matrix $\bm{A}$ appropriately. 
As shown in Eq. (\ref{eq:abar}), matrix $\bm{A}$ is the main factor to determine the time constant of memory retention. 
When the matrix $\bm{A}$ is learned as a neural network parameter, the time constant of memory retention 
is learned to minimize the error function. 
If designers want to assign the memory time constant arbitrarily, matrix $\bm{A}$ can be given as a fixed value. 
Although there is a trade-off between the length of memory retention and 
the frequency bands of extractable features, 
it is possible to balance both by comprehensively setting high and low time constants 
since the eigenvalues of matrix $\bm{A}$ can be set multiple times.

\section{Evaluation}

\subsection{Robot Configuration}
\begin{figure}[tb]
  \centering
  \includegraphics[width=4.2cm]{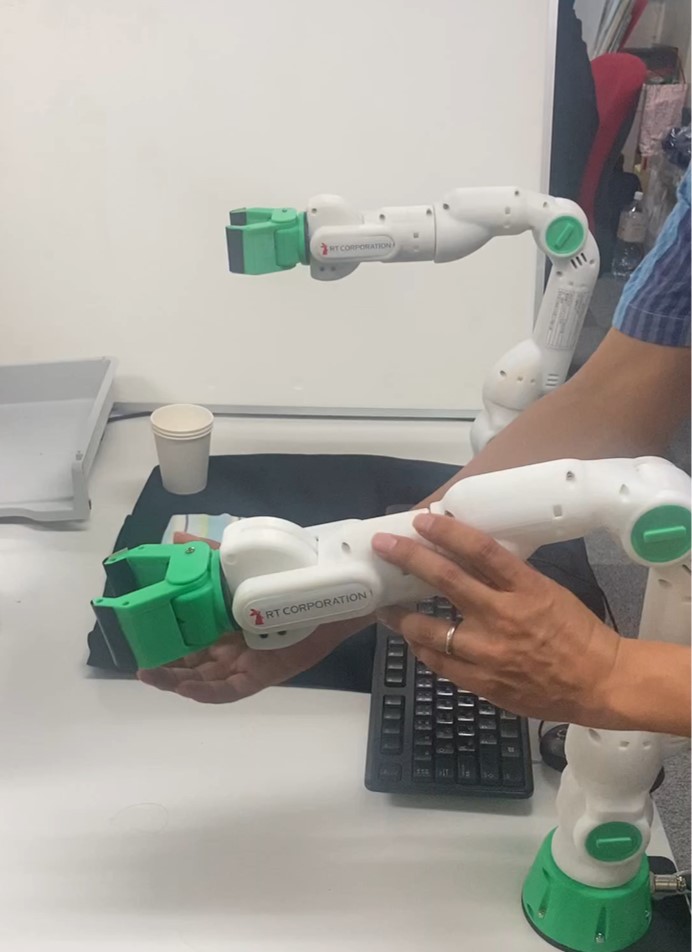}
  \caption{Training using bilateral control system.}
  \label{fig:bilateral}
\end{figure}

Training data was collected through bilateral control of an 8-degree-of-freedom robot, as shown in Fig.~\ref{fig:bilateral}.  
The performance of imitation learning was evaluated using the follower robot. 
The robot employed in this study was the CRANE-X7 by RT Corporation, which features a 7-axis manipulator 
with a 1-axis gripper at the end. A single control PC was used to calculate inputs for both robots, performing 
bilateral control for each axis in joint space~\cite{sasagawa2020imitation}. 
The control PC communicated with a dedicated machine learning server equipped with a GPU via UDP protocol, 
operating at a 100 ms cycle, while the control loop maintained a 2 ms cycle time.
In the neural network architecture, convolutional layers with a kernel size of 4 were employed. 
The model was trained using mean squared error (MSE) as the loss function, and optimization was performed 
using the AdamW algorithm with a learning rate of 1e-3. 

\subsection{Evaluation Method}
This paper presents offline and online evaluations of motion imitation learning, respectively.
The offline test aimed to quantitatively evaluate the model's trajectory prediction performance considering long-term dependencies. 
Motions were generated using a model trained with training data obtained from the experimental system, and errors in the predicted 
output were evaluated when test data from the experimental system was given as input.
The model was evaluated on three tasks of increasing complexity: simple up-and-down movements, cup placement after grasping, and case loading.
The up-and-down motion was first repeated for 10 seconds.  
Next, the motion was repeated twice and then stopped.
In the former trial, the next trajectory can be predicted based only on the short-term memory, while in the latter trial, long-term memory 
is required to predict the stopping after repeated movements.
In addition, cup placing and case loading were tested as tasks simulating actual work. 
In the former task, the robot grasped an empty 7-cm-diameter, 9-cm-high cup placed in a fixed position and evaluated whether 
the cup was placed in a specified 12-cm-square area without tipping it over.
The latter evaluated whether the subject could grasp a paper box 21 cm long, 3 cm wide, and 8 cm high, and place it in an upright position in a 22-cm-width case.
Since the clearances between the case and the box was 1 cm, the box often contacted the case during training, requiring the instructor to adjust the position accordingly.
Therefore, the task had training data with a higher variability of motion.
The robot in this study was not equipped with a vision system and reproduced a situation in which the position of the object was almost 
identified by placing the object in position with an error of 5 mm or less.
Hyperparameters of LSTM and Transformer were determined with reference to the studies of 
 \cite{sasagawa2020imitation,kobayashi2024ilbit}, respectively, which have the closest control system configuration.
For all the tasks, the number of trials for training and test data were 18 and 6, respectively.
\subsection{Offline Evaluation}
The errors of each model are compared in the Table~\ref{table:rmse}.
Here, in order to have the same evaluation as that used for the error function, the error is based on the integration of information in different units, angle (rad) and torque (Nm).
The Transformer model with full period input consistently demonstrated superior performance across all tasks, 
whereas the Transformer with short period input showed performance degradation due to the unavailability of past history information.
This is supported by the fact that the degradation of the short period input transformer is relatively low for repetitive up-and-down movements, which do not require historical information.
Although the root mean squared error (RMSE) of Mamba was not as small as that of the Transformer with full period input, it was confirmed that a similar performance can be obtained with only short period input and contextual state variables.

\begin{table}
  \caption{Comparison of estimation error.}\label{table:rmse}
  \small
  \begin{tabular}{l|cccc}
    \hline
\\[-0.5em]
    Model & \!Up\&Down\! & \!Up\&Down\! & \!Cup & \!Case  \\
             & \!(repetitive)\! & (twice) & \!\!Placing & \!\!Loading \\[0.5em]
    \hline
\\[-0.5em]
    LSTM   &\!\!0.449      & \!\!\!\! 0.418        & $\!\!\!\!\!\! 0.812$         & $\!\!\!\!\!\! 0.916$\\
              &$\ \pm$0.350 & $\ \pm$0.275\!\! & $\ \pm$0.346 & $\ \pm$0.456\\[0.5em]
    Transformer\!\!&\!\!\!\! 0.132        & \!\!\!\! 0.387         & $\!\!\!\!\!\! 0.135$         & $\!\!\!\!\!\! 0.137$ \\
    (w=T)        &$\ \pm$0.054\!\! & $\ \pm$0.304 & $\ \pm$0.053 &  $\ \pm$0.056 \\[0.5em]
    Transformer\!\!&\!\!\!\! 0.330        & \!\!\!\! 0.863         & $\!\!\!\!\!\! 0.296$          & $\!\!\!\!\!\! 1.09$ \\
    (w=20)       &$\ \pm$0.264\!\! & $\ \pm$0.501 & $\ \pm$0.159 & $\ \pm$ 0.34 \\[0.5em]
    Mamba    &\!\!\!\! 0.277  & \!\!\!\! 0.164     &     \!\!\!\!\!\!0.262           &$\!\!\!\!\!\! 0.230$ \\
    (w=20)    &$\ \pm$0.236\!\!    & $\ \pm$0.142  &  $\ \pm$0.155  & $\ \pm$0.029\\[0.5em]
    \hline
  \end{tabular}
\end{table}

\begin{figure}[tb]
  \centering
  \includegraphics[width=8.4cm]{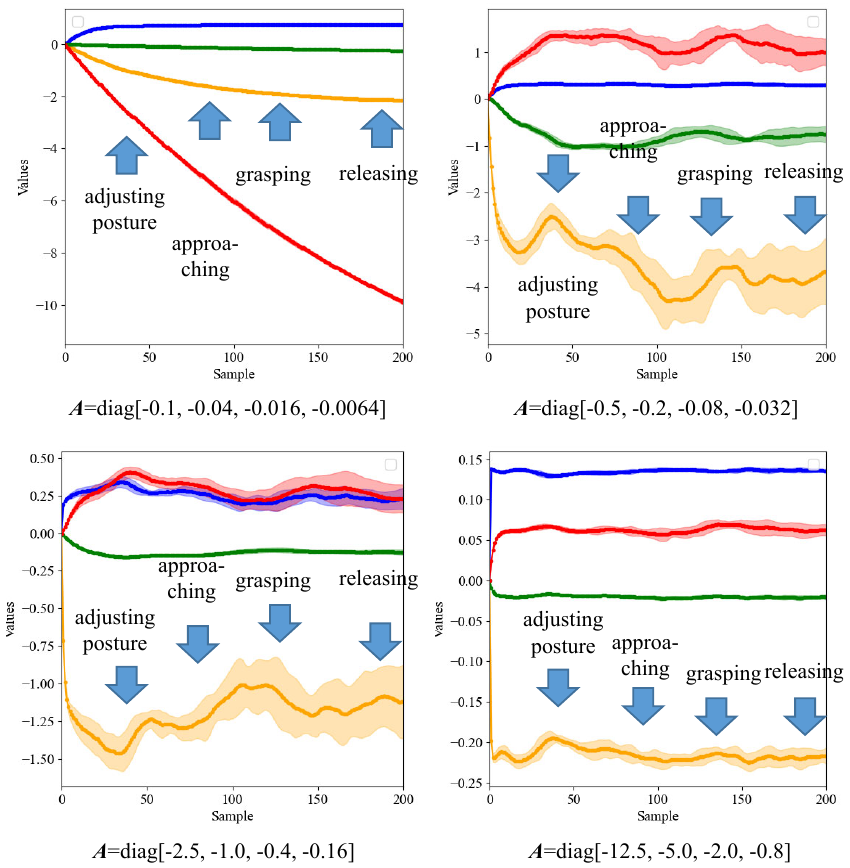}\\
  \scriptsize{(a) Matrix $\bm{A}$ fixed }\\
  \includegraphics[width=4.6cm]{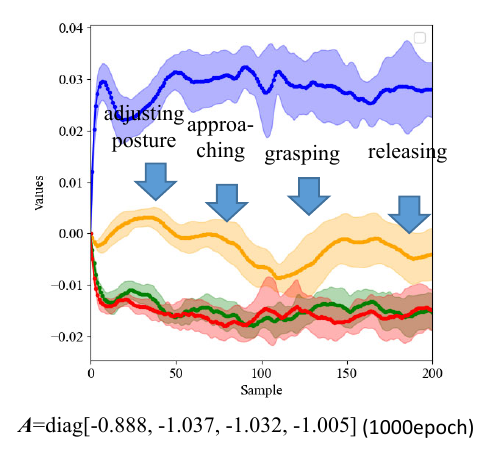}\\
  \scriptsize{(b) Matrix $\bm{A}$ trained }
  \caption{Time responses of variables in state space.}
  \label{fig;timeresponses}
\end{figure}

Next, differences were examined between learning the diagonal elements of matrix $\bm{A}$ with a neural network and keeping them fixed. 
Fig.~\ref{fig;timeresponses}(a) shows the time responses of the state variable $\bm{h}_t$ for Mamba with the length of the input $w=20$ under different fixed $\bm{A}$ matrices. 
To confirm whether feature values can be obtained from state variables as work progresses, arrows were added to indicate timings of work progression 
such as posture alignment, approaching the cup, grasping, and releasing.
In all examples, the diagonal elements were given values multiplied by 0.4 from the minimum value to ensure uniform variation between axes of $\bm{A}$.
For the autoencoder to perform sufficiently well, it is desirable for the SSM, which corresponds to latent variables, to be dispersed throughout the space. 
Also, to extract task features with high performance, it is desirable for state variables to fluctuate in conjunction with characteristic input signals.

When the elements of $\bm{A}$ were large (-0.1 at minimum), it was confirmed that the SSM response values varied greatly over time. 
However, there was significant bias in the spatial distribution, and it was confirmed that the slope of the state variables was almost uniform, 
with changes dependent on input being small compared to the overall response values.
When the elements of $\bm{A}$ were small (-12.5 at minimum), the SSM values did not change significantly depending on time. 
In both cases of large and small $\bm{A}$, the variance of the state variables became small. 
This means that the model did not significantly reflect the differences in responses between trials.
In the results where the minimum values of $\bm{A}$ elements were -0.5 and -2.5, i.e. the elements of $\bm{A}$ were not too small or large,  
the state variables were widely distributed in space and changed as the work progressed. 
In sum, to extract low-dimensional features from input information into the state space, the elements of matrix  
$\bm{A}$ need to be given appropriately to achieve suitable time constants. 

\begin{figure}[tb]
  \centering
  \includegraphics[width=8.2cm]{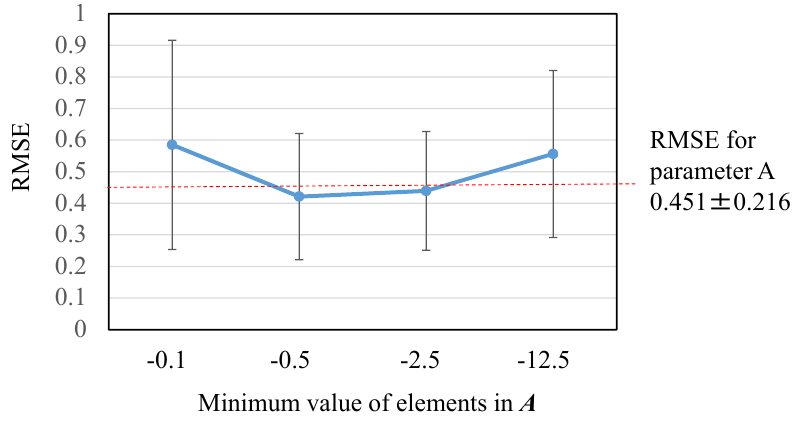}
  \caption{RMSE of fixed A matrices.}
  \label{fig:fixeda}
\end{figure}

Fig.~\ref{fig;timeresponses}(b) shows the results when matrix $\bm{A}$ was learned as parameters of the NN. 
While the variance of all state variables increased, the diagonal elements of $\bm{A}$ did not deviate significantly 
from their initial values -1, and some state variables showed similar behavior, indicating a redundant configuration.
Fig.~\ref{fig:fixeda} compares the RMSE when the diagonal elements of matrix  $\bm{A}$ are given as fixed values.
When matrix  $\bm{A}$ is too large or too small, the prediction accuracy of robot motion deteriorates, but when the elements of 
 $\bm{A}$ are set within an appropriate range, high prediction accuracy is obtained.
The red dotted line represents the average prediction accuracy of 0.451$\pm$0.216, when the diagonal elements of matrix 
$\bm{A}$ were learned as parameters of the neural network. When appropriate fixed values were given, the prediction accuracy 
was better compared to when adjusted by machine learning.
This result suggests the advantage of giving matrix $\bm{A}$ as fixed values rather than learning it with a neural network, 
especially when the amount of training data is limited, because it results in a model with lower degrees of freedom.

\begin{figure}[tb]
  \centering
  \includegraphics[width=8.2cm]{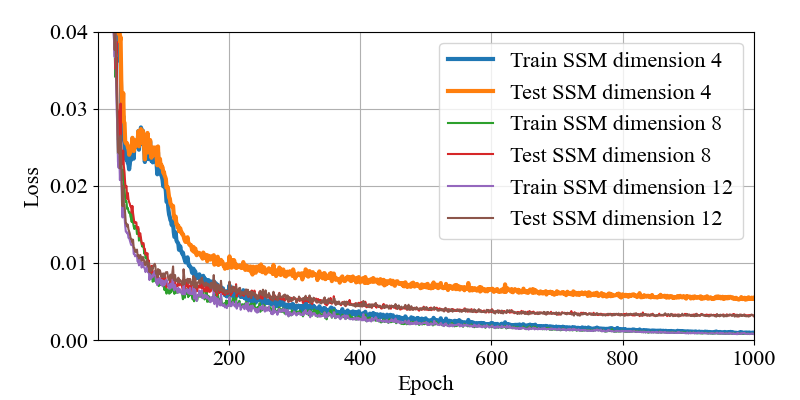}\\
  \caption{Loss curves for different SSM dimensionalities.}
  \label{fig:losscurve}
\end{figure}

Fig.~\ref{fig:losscurve} shows the change in loss curves depending on the dimensionality of SSM. 
As the dimensionality increases, the learning speed increases and the test loss decreases, but beyond a certain dimension, 
the improvement becomes negligible. This result indicates that the SSM can be used for imitation learning with 
little performance degradation even when the dimensionality is reduced to some extent. 

\begin{figure}[tb]
  \centering
  \includegraphics[width=8.2cm]{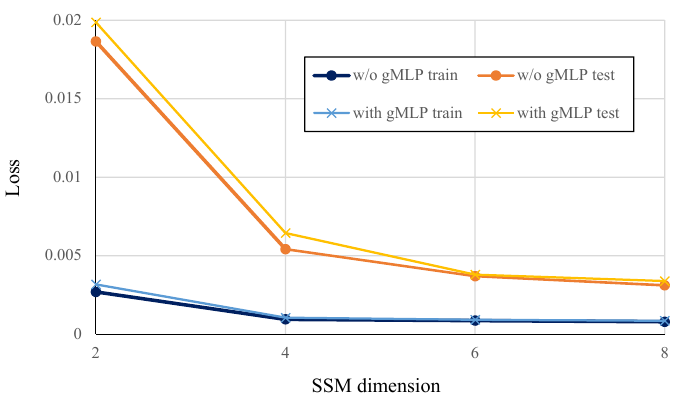}\\
  \caption{Performance comparison with and without gMLP.}
  \label{fig:withandwithout}
\end{figure}

Fig.~\ref{fig:withandwithout} shows the difference in performance with and without gMLP. 
For all results, training data from 16 trials of Cup Placing and test data from 8 trials were used for evaluation. 
There was no significant difference between with and without gMLP models, when the SSM dimension is 6 or higher,  
while, for results where the SSM dimension is 4 or lower, models without gMLP demonstrated better results.
In current robotics, imitation learning is applied to relatively simple tasks. 
When the order of the model is reduced for such tasks, performance may improve by excluding gMLP.


\subsection{On-line evaluation}
Table~\ref{table:sr} shows the success rates when each task was performed 20 times on the actual machine.

To mitigate the substantial oscillations observed in the Transformer's output
a first-order low-pass filter with a time constant of 0.3 seconds 
was applied to the output to suppress vibrations, and the filtered values were given to the robot as command values to improve 
the task success rate. As a result of this adjustment, the Transformer with full period input showed the smallest output RMSE and demonstrated 
high success rates in actual machine tasks. However, a certain percentage of failures occurred in both Cup Placing and Case Loading tasks. 
The main factors for failure were dropping of grasped objects due to vibration and unexpected contact with the environment 
that was not present during training. These issues were more pronounced in the Transformer with short period input, resulting 
in further decreased success rates. 

On the other hand, when performing tasks using Mamba, there was almost no vibration, resulting in a 100\% success rate for 
Cup Placing, which has a large tolerance for position deviations. In Case Loading, there were two instances where improper contact 
with the case prevented accurate placement in the desired location, but the overall success rate was higher than that of the Transformer. 
This is thought to be due to the minimal vibration in Mamba's output, despite its RMSE being larger than the Transformer's.
Transformers are known to generally have weak inductive bias, which often brings positive effects. 
However, in robot motion generation, this characteristic does not guarantee output continuity and tends to cause vibrations. 
Mamba does not generate such vibrations because the SSM has a structure that integrates state variables, 
incorporating the regularity of continuous output changes into the model. 
This functions as an inductive bias, enabling efficient learning from data. 
In robot trajectory generation, continuity based on dynamic constraints is crucial, so the inductive bias provided by SSMs 
like Mamba can have a positive effect on robot imitation learning.

Furthermore, to evaluate real-time motion generation capabilities, Mamba's computation time was reduced to 25ms for experiment. 
The success rate is shown in the 4x speed row of Table~\ref{table:sr}. 
As a result of the sampling time being reduced to one-fourth of the original, motions were generated four times faster. 
Although this acceleration increased vibrations due to dynamics inconsistencies and raised the failure rate, 
the overall decrease in success rate was minimal, confirming Mamba's real-time adaptability and stability.

\begin{table}
  \centering
  \caption{Success rate in experiment.}\label{table:sr}
  \small
  \begin{tabular}{l|cccc}
    \hline
\\[-0.7em]
    Model &  Cup Placing & Storing  \\[0.3em]
    \hline 
\\[-0.5em]
    Transformer (w=T) & 16/20 & 13/20 \\[0.2em]
    Transformer (w=20)& 0/20 & 4/20 \\[0.2em]
    Mamba (w=20)& 20/20 & 18/20 \\[0.2em]
    Mamba (w=20,4x speed)& 19/20 & 14/20 \\[0.3em]
    \hline
  \end{tabular}
\end{table}

\begin{figure*}[tb]
  \centering
  \includegraphics[width=17.2cm]{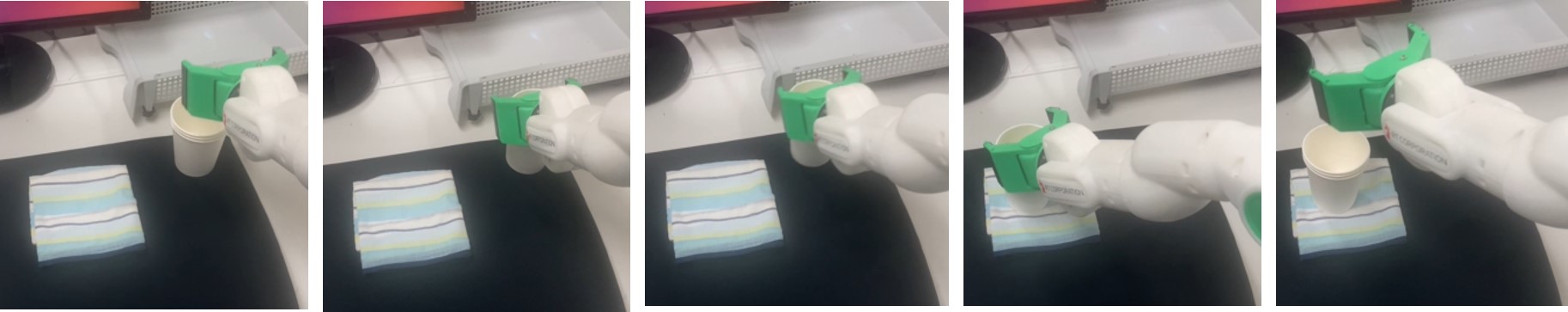}
  \caption{Cup placing task.}
  \label{fig:expcup}
  \vspace{5mm}
  \includegraphics[width=17.2cm]{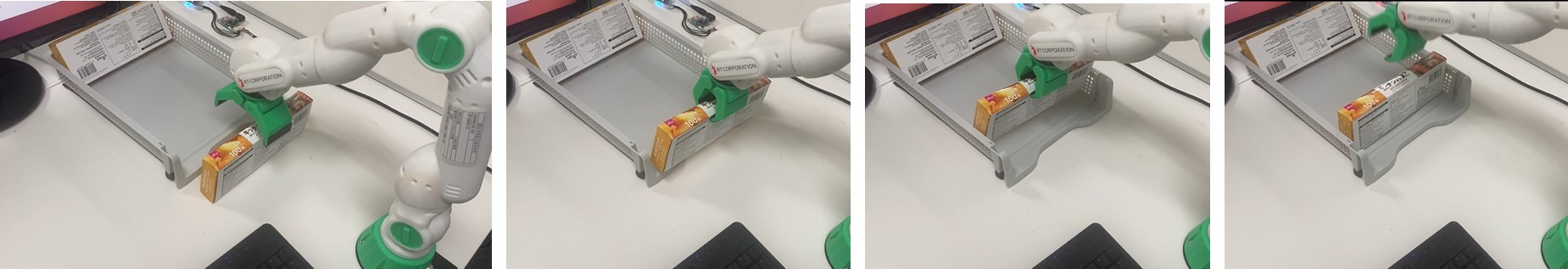}
  \caption{Case loading task.}
  \label{fig:expbox}
\end{figure*}

\section{Limitation}
The model proposed in this research generates motion based on state variables and 
short-term sequences rather than using all sequences as input. 
While this approach effectively reduces the network structure and the scale of data 
that needs to be retained, it may impact the Selection Mechanism. 
When all sequences are inputted, the Selection Mechanism captures long-term dependencies 
across all inputs and filters out unnecessary information. 
In the proposed model, information is filtered based on the values of state variables given 
as past context and the input from a limited period. 
The experimental results in this paper suggest that this does not hinder task performance; 
however, the impact may vary depending on the task being applied, necessitating further investigation.

This study presents a modified version of Mamba, excluding gMLP, which has demonstrated high performance in experiment. 
However, it remains unclear to what extent the exclusion 
of gMLP offers advantages. The tasks addressed in this research are relatively simple and can be realized 
within a context similar to System 1 of dual-process theory~\cite{frankish2010dual}, which requires a less complicated reasoning. 
For tasks requiring more advanced reasoning, akin to System 2, it is likely that gMLP could be necessary. 
Therefore, it is essential to investigate the relationship between the amount of training data, the anticipated 
modalities, the tasks to be imitated, and the appropriate model design, for developing design principles. 

\section{Conclusion}
This study demonstrates that Mamba offers significant 
advantages for robotic imitation learning. 
By reducing the dimensionality of the state space, Mamba functions effectively as an autoencoder, 
compressing past contextual information into low-dimensional state variables. 
A model is proposed that generates motion based on the contexual state variables and short period input sequences.
This approach reduces both the complexity of 
the network structure and the amount of data storage required, compared to the model that uses full-period inputs.
This characteristic enables efficient learning and robust performance in real-time motion generation. 

The experiments revealed that, despite a higher RMSE compared to Transformers, 
Mamba consistently achieved higher success rates in robotic tasks, owing to its ability to suppress 
vibrations and maintain continuity in output trajectories. This highlights the importance of inductive 
bias in robotic systems, where the integration of state variables within a low-dimensional SSM 
framework contributes to more reliable and stable task execution.

Furthermore, Mamba's ability to generate real-time motions with reduced computation time underscores 
its potential for deployment in dynamic environments where rapid and adaptive responses are critical. 
These findings suggest that SSMs like Mamba, with their built-in structural advantages, 
are well-suited for advancing the capabilities of robot imitation learning, providing a strong foundation for future developments in this field.

\small
\bibliography{biblist} 
\bibliographystyle{IEEEtran} 

\end{document}